\documentclass[sigconf]{acmart}
\usepackage{makecell}
\usepackage{graphicx}
\usepackage{booktabs}
\usepackage{subcaption}
\usepackage{xcolor}
\usepackage{multirow}
\usepackage{url}

\usepackage{textcomp}
\usepackage[titlenumbered,ruled,noend]{algorithm2e}
\usepackage{mathtools}
\usepackage{bbold}
\usepackage{pgfplots}
\usepackage{tikz}

\DeclarePairedDelimiter\floor{\lfloor}{\rfloor}

\AtBeginDocument{%
  \providecommand\BibTeX{{%
    \normalfont B\kern-0.5em{\scshape i\kern-0.25em b}\kern-0.8em\TeX}}}

\setcopyright{acmcopyright}
\copyrightyear{2018}
\acmYear{2018}
\acmDOI{10.1145/1122445.1122456}

\acmConference[Woodstock '18]{Woodstock '18: ACM Symposium on Neural
  Gaze Detection}{June 03--05, 2018}{Woodstock, NY}
\acmBooktitle{Woodstock '18: ACM Symposium on Neural Gaze Detection,
  June 03--05, 2018, Woodstock, NY}
\acmPrice{15.00}
\acmISBN{978-1-4503-XXXX-X/18/06}

\begin{document}
\title{Detecting Anomalies Through Contrast in Heterogeneous Data}

\author{Debanjan Datta}
\email{ddatta@vt.edu}
\affiliation{%
  \institution{Discovery Analytics Center, Virginia Tech}
  \city{Arlington}
  \state{VA}
  \country{USA}
  \postcode{22203}
}

\author{Sathappan Muthiah}
\email{sathap1@vt.edu}
\affiliation{%
  \institution{Discovery Analytics Center, Virginia Tech}
  \city{Arlington}
  \state{VA}
  \country{USA}
  \postcode{22203}
}

\author{Naren Ramakrishnan}
\email{naren@vt.edu}
\affiliation{%
  \institution{Discovery Analytics Center, Virginia Tech}
  \city{Arlington}
  \state{VA}
  \country{USA}
  \postcode{22203}
}


\begin{abstract}
Detecting anomalies has been a fundamental approach in detecting potentially fraudulent activities. 
Tasked with detection of illegal timber trade that threatens ecosystems and economies and association with other illegal activities, we formulate our problem as one of anomaly detection.
Among other challenges annotations are unavailable for our large-scale trade data with heterogeneous features (categorical and continuous), that can assist in building automated systems to detect fraudulent transactions.
Modelling the task as unsupervised anomaly detection, we propose a novel model \textit{\textbf{C}ontrastive  Learning based \textbf{H}eterogeneous \textbf{A}nomaly \textbf{D}etector} to address shortcomings of prior models.
Our model uses an asymmetric autoencoder that can effectively handle large arity categorical variables, but avoids assumptions about structure of data in low-dimensional latent space and is robust to changes to hyperparameters.
The likelihood of data is approximated through an estimator network, which is jointly trained with the autoencoder, using negative sampling.
Further the details and intuition for an effective negative sample generation approach for heterogeneous data are outlined. 
We provide a qualitative study to showcase the effectiveness of our model in detecting anomalies in timber trade.
\end{abstract}



\maketitle
\section{Introduction}{
Illegal trade in ecologically vulnerable timber is detrimental to not just ever-diminishing biodiversity, but also adversely impacts developing economies and is linked to illicit financial flows~\cite{lawson2010illegal}.
Mislabelling or mis-classification of shipments has been investigated in timber trade ~\cite{wiedenhoeft2019fraud}, and poses a significant problem.
With the US being one of the largest importers of timber products, there has been sustained efforts in detecting and thwarting such shipments.
Yet an automated system to detect potentially suspicious trades is not readily available due to hurdles in policy and inter-institution (governmental and conservationist) collaboration, absence of annotated data and prohibitively high cost of human annotation for trade data. 
Our work is part of an effort which is a collaboration with multiple conservationist agencies who are working to address such shipments.

Following prior approaches towards the task of fraud detection, we model the task as unsupervised anomaly detection. 
However the existing approaches for anomaly detection in  mixed or heterogeneous  data have a few drawbacks, given the nature of our target data. 
Firstly, they are not designed for categorical variables with high arity (cardinality).
Secondly, model designs often have sensitive hyperparameters such as cluster numbers which are difficult to determine apriori in a true unsupervised scenario. 
Thirdly, there is often a reliance on various distributional assumptions, which may be not be satisfied by arbitrarily complex real world data.
It is imperative to overcome these impediments for our task, and make the model free of such limiting assumptions so as to create a automated fraud detection framework applicable to complex real-world problems like timber trade.

\begin{figure}[!tp]
\centering
\includegraphics[width=\columnwidth]{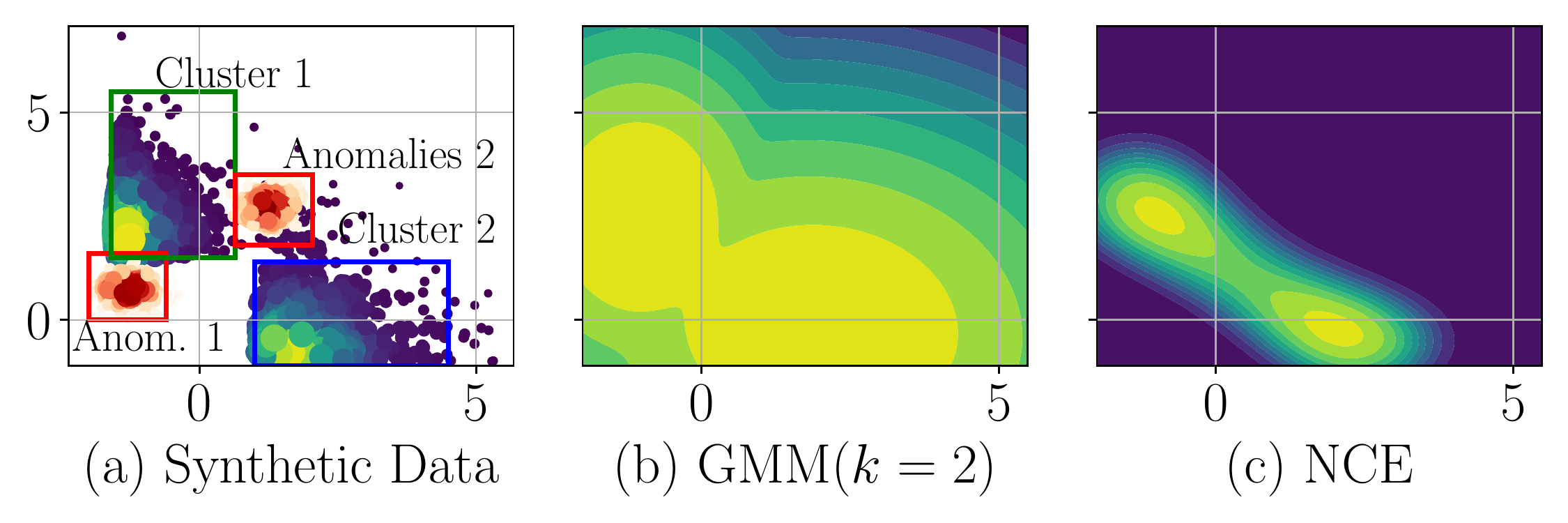}
\caption{(a) shows an example scenario where data clusters are of arbitrary shape. (b) show the density as learned by GMM and (c) the density learned through contrast using Noise Contrastive Estimation. We can clearly see that the Gaussian assumption made by GMM does not generalize well to such data and thus can significantly affect anomaly detection performance}
\label{fig:concept}
\end{figure}

We propose a novel model, Contrastive Learning based Heterogeneous Anomaly Detector~(CHAD) that attempts to alleviate these problems.
The model uses an autoencoder based architecture to obtain a low-dimensional latent representation of data, with additional network features to handle potentially high arities in categorical variables along with  NCE~\cite{gutmann2010noise} based approach for detecting anomalies. Our contributions are: (1) \textbf{Direct likelihood estimation}, unlike prior approaches we do not make any distributional assumptions, instead rely on contrastive learning for flexibility and generalizability; (2) \textbf{Effective negative sampling} approach for heterogeneous data; (3) \textbf{Robustness} to changes in hyper-parameters and finally, (4) \textbf{Improved (or competitive) performance} over existing state-of-the-art across multiple open source datasets.
}

\section{Related Works}{
Unsupervised anomaly detection methods for general data have been discussed in detail in surveys such  \cite{chandola2009anomaly}, \cite{goldstein2016comparative} and \cite{chalapathy2019deep}.
Further, application specific contexts like fraud detection~\cite{abdallah2016fraud} and intrusion detection~\cite{khraisat2019survey}.
We organize our related works as follows:

Traditional approaches approaches include Kernel Density Estimation~\cite{latecki2007outlier}, Principal Component Analysis and Robust PCA~\cite{xu2010robust}. 
Some other notable methods are Local Outlier Factor\cite{breunig2000lof} and  Isolation Forest~\cite{liu2008isolation}. A key challenge with such methods is that they are not scalable, and many of them are not effective for sparse high dimensional input features - which are important in our use case.

\textbf{Clustering} based anomaly detection approaches assume \textit{normal} or expected data to posses proximity to each other along with some latent underlying structure, which has been can be utilized to detect outliers. 
Recent approaches like Deep Embedded Clustering (DEC)~\cite{xie2016unsupervised} and Deep Clustering Network (DCN)~\cite{yang2017towards} are relevant here. Mixture models like 
DAGMM~\cite{zong2018deep} which combines a deep autoencoder with Gaussian mixture model to perform anomaly detection can also be considered under clustering methods.
An interesting aspect of this approach is a very low dimensional latent representation is augmented with sample reconstruction errors.

\textbf{Autoencoders} utilize a reconstruction-based approach relying on the assumption that anomalous data cannot be represented and reconstructed accurately by a model trained on \textit{normal} data.
Autoencoders and its variants such as denoising autoencoders have been used for anomaly detection in works such as those presented in \cite{sakurada2014ae}),~\cite{tagawa2015structured} and \cite{zhou2017anomaly}, while \citeauthor{chen2017outlier}~(\citeyear{chen2017outlier}) use an 
autoencoder ensemble.

\textbf{One-class classification} based approaches like OCSVM~\cite{scholkopf2000support} separates the normal data from the anomalies using a hyper-plane of maximal distance from the origin.
While another model, the Support Vector Data Description~\cite{tax2004support} attempts to find the smallest hyper-sphere that contains all normal data. 
Both methods use a hyperparamter $\nu$ that helps define the boundary of the hyper-plane or hyper-sphere. 
Similarly, two recent works employ deep learning along with a one-class objective for anomaly detection.
DeepSVDD~\cite{ruff2018deep} performs anomaly detection combining deep convolutional network  as feature extractor and one-class classification based objective. 
Oneclass-NN~\cite{chalapathy2018anomaly} uses a similar formulation, albeit using a hyper-plane instead of a hyper-sphere, and an alternating minimization based training approach. 
Finally, two recent methods specific to mutivariate categorical features are \textit{APE}~\cite{chen2016entity} and \textit{MEAD}~\cite{datta2020detecting}. 
Though \textit{MEAD} uses a trade dataset, it considers only categorical features. Hence is not applicable for our study.
Both represent entities in a common embedding space and interactions between them are computed to obtain the likelihood of a record. 

Most of the above mentioned approaches make critical assumptions about the data such as cluster numbers, distributional family, etc. 
These are often specific to a particular kind of data and also it is difficult to obtain necessary domain knowledge to be able to make such choices.
In the following section we show a simple conceptual example denoting the disadvantages of making such assumptions about data and propose an alternative non-parametric density estimation method.
}

\section{Conceptual Motivation}\label{sec:concept}{
To detect anomalies, a standard approach is to perform clustering that assume the shape of data is elliptical or spherical in space. 
We provide conceptual proof using a simple scenario with 2-dimensional data as shown in Figure~\ref{fig:concept} to showcase the limitations of such approaches.
Two clusters ($C_1$ and $C_2)$ of random data points points are drawn from an arbitrary choice of two independent bi-variate Gamma distributions($G_1$ and $G_2$).
Another two sets of points, which are treated as anomalies are generated using Normal distributions denoted as $N_1$ and $N_2$.
Let us model the data using Gaussian Mixture Model (GMM) with $k=2$ components and also with K-Means with $k=2$ .
We estimate the anomalousness of a point as the distance from assigned cluster center for the K-Means model and as the likelihood of a data point conditioned on the estimated model parameters for GMM (similar to DAGMM proposed in~\cite{zong2018deep}. 

Scoring the points, we find the average precision (area under the Precision-Recall curve) for GMM to be below $0.5$ and 0.86 for K-Means. Here for K-means with $k=2$ the assumed number of clusters happens to match with true data and hence shows a reasonably good performance. However such knowledge is not readily available in timber trade data of interest to us. The average precision drops significantly to $0.2$ when changing the number of clusters to $k=1$. Such an assumption is made in cases like deepSVDD~\cite{ruff2018deep}.
There exists literature on estimating the optimal number of clusters for a given dataset such as X-means~\cite{pelleg2000x} or Bayesian GMM~\cite{rasmussen2000infinite}.
However bayesian inference is known to suffer from scalability issues in complex (high dimensional) and big data models. We also note, for a simple scenario like the one considered above, determining the appropriate number of clusters through tests such as silhouette analysis would be easy, but for sparse high dimensional data such analysis is prone to errors.

We can depart from such limiting assumptions if we could approximate a \textit{contrastive} distribution, whose density is high only in regions where the \textit{normal} data is absent. Provided $G_1$ and $G_2$ is revealed to us, this can be easily done by sampling from an uniform distribution $U$ over the valid data domain such that $P_{G_1}(u)< \epsilon$ and $P_{G_2}(u)< \epsilon$. 
Sampling from this distribution, and from the \textit{normal} data, it is possible to create a discriminator function that can predict which region of space a point belongs, or in other words can estimate the probability of a point belonging to region of \textit{normal} data following the concept of NCE~\cite{gutmann2010noise}. Such a model achieves an average precision of $0.92$. 

This shows a model that estimates the density of \textit{normal} or expected data using a contrastive distribution can be effectively used for anomaly detection. 
The key challenge is that in real world data neither is analytical form of data distribution exactly known, nor is it a trivial task to estimate such a contrastive distribution. 
In the upcoming sections we detail how to overcome this problem and estimate the contrastive distribution through our model.
}

\section{Model Architecture}\label{sec:model}{
\begin{figure}[t]
\centering
\includegraphics[width=0.950\columnwidth]{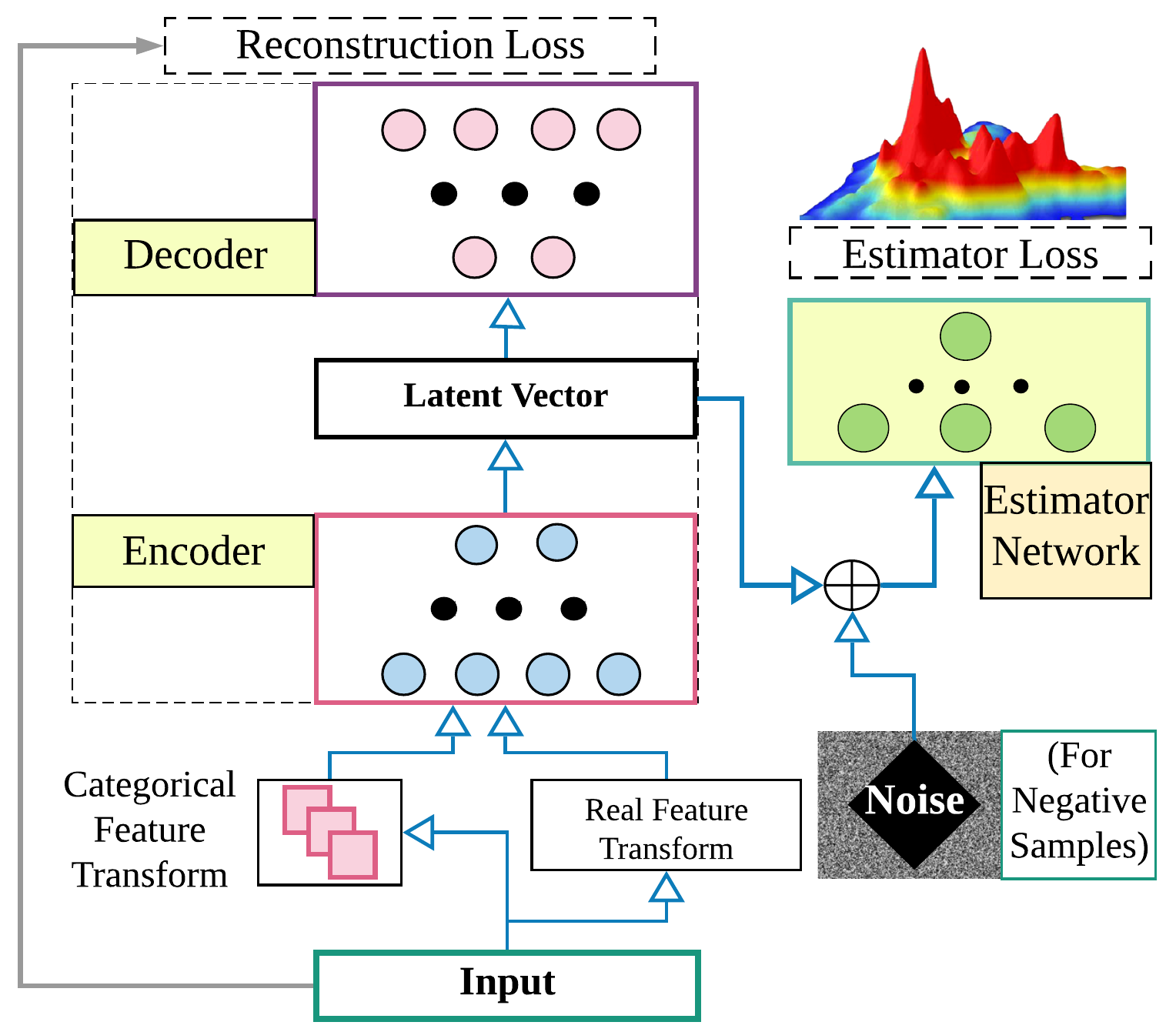}
\caption{Proposed architecture of our model Contrastive Learning based Heterogeneous Anomaly Detector(CHAD)}
\label{fig:model}
\vspace{-1em}
\end{figure}

Trade data contains heterogeneous features with high arity categorical features such as Harmonized Schedule Codes, ports of lading and unlading, origin and destination along with numerical features such as weight, volume and worth.
\textbf{Problem Statement}: Given a dataset $\mathcal{D}$ containing heterogeneous attributes which are assumed to be clean, learn a model $\mathcal{M}(\theta)$ that can predict $P(x |\mathcal{\theta})$, the likelihood of a test record being drawn from the underlying data distribution. 
The test records with likelihood below a user-defined threshold are deemed anomalous.

While embedding based architectures have been used in prior works, evidence shows deep autoencoders preserve richer information in a reduced dimension compared to shallow or linear autoencoders~\cite{hinton2006reducing}.
However autoencoder based approaches presented in prior works have a drawback since they use reconstruction error as anomaly scores. 
This is because if the autoencoder generalizes too well - it can lead to anomalous samples having low reconstruction error~\cite{spigler2019denoising}.
In line with these observations we make use of autoencoders for modeling our data but instead of using reconstruction loss as anomalousness score we use the learned latent representation to build a classifier that learns to discriminate between observed data and random noise. The classifier softmax score is then used as the score in our model. Details of our autoencoder architecture is presented next.

\subsection{Asymmetric Autoencoder}{
Autoencoders generally have symmetric architectures for the encoder and the decoder, however we propose the use of an asymmetric architecture.
The decoder is a fully connected dense neural network with dropout.
An additional embedding (or linear transform) layer is added prior to the fully connected layers of the encoder.
This layer is designed to handle the different categorical variables (domains) with potentially high arities.

We can also include a similar (non-)linear transform for the set of all continuous features if the dimensionality is deemed high.
This requires decision on part of the practitioner, but we find it helps to add a transform where dimensionality is over 32.

Formally, let $\mathbf{x} \in \mathcal{D}$ be an input. $\mathbf{x}$ is a multidimensional variable, with $d$ features.
Let $x_1,x_2...x_k$ be categorical features, and $x_{k+1}...x_d$ be continuous features.
$x_{k+1}...x_d$ can be considered as a multivariate feature, denoted as $x_r \in R^{d-k}$. 
Let $f_{i}(.)$ denote the linear transform or the identity transform, for $i^{th}$ categorical field. 
Let $g(.)$ denote transformation on $x_r$. Let $x_t$ denote the transformed input, which is input to the fully connected layers of the encoder, as shown in Eq.~\ref{eq:x_e}.
Both the encoder and decoder utilize dropout and a non-linear activation function(tanh), except for the final output where sigmoid is used.
\begin{equation}
   x_t = f_1(x_1) \oplus f_2(x_2) ... \oplus f_k(x_k) \oplus g(x_r)
   \label{eq:x_e}
\end{equation}
The autoencoder is designed to optimize reconstruction of the input vector, which is an auxiliary task in our case since we are interested in the latent representation. 
We choose Mean Squared Error for the reconstruction Loss, denoted as $\mathcal{L_R}$ as shown in Eq.~\ref{eq:loss1}. 
\begin{equation}
    \mathcal{L_R} = \frac{1}{N}\sum_{i}(\mathbf{x}-\mathbf{\hat{x}})^2
    \label{eq:loss1}
\end{equation}

While it is possible to have separate losses for categorical and numerical features, we do not find any significant performance difference in doing so.
In the next section, we detail out how the latent representation is used in the discriminator module to identify data density.
}

\subsection{Density Estimation Network}{
The core idea here is to learn the density of data through comparison or contrast with an artificially generated noise distribution following~\citeauthor{gutmann2010noise}~(\citeyear{gutmann2010noise}).
Let 1 and 0 be the class labels for the normal data and noise respectively. 
Let us also start with an assumption that $P(C=1)=P(C=0)=0.5$ and we denote the empirical distribution of noise as $p_n(.)$. 
We model the density of data as $p_d(.;\theta)$, with $\theta$ being model parameters. That is, $P(x|C=1,\theta)=p_d(x;\theta)$ and $P(x|C=0)=p_n(x)$. Therefore,
\begin{equation}
    P(C=1|x;\theta) = \frac{p_d(x;\theta)}{p_d(x;\theta) + p_n(x)} = f(x;\theta)
\end{equation}
Here $P(C=1|x;\theta)$ is the posterior distribution of class label 1, and $f(x;\theta)$ is the approximate estimation function.
A MLP can be utilized as functional approximator for $f(x;\theta)$ without loss of generality. 
We wish to maximize the log likelihood, to find the set of parameters $\hat{\theta}$ for the model that maximize the likelihood of the training data. 
\begin{equation}
\small
\begin{aligned}
    \mathcal{L}(\hat{\theta}) =   \arg \underset{\theta} \max \sum_{i} C_i ln(P(C=1|x_i;\theta) \\ + (1-C_i) ln(P(C=0|x_i;\theta)
    \\
    \approx \sum_{i}ln(f(x_i;\theta)) + ln\big(1 - \frac{1}{|k|}\sum_{k}(f(z_k;\theta)\big)
\end{aligned}
\label{eq:est1}
\end{equation}   
In the second line of Eq.~\ref{eq:est1}, $z_k$ refers to the $k$th negative sample drawn for each observed instance $x_i$. This is because we do not have access to actual samples from class 0.

Now, we can replace $x_i$ with $x_e$ in Eq.~\ref{eq:est1}, where $x_e = \mathbf{f_{enc}}(x_i)$ and $x_i \in \mathcal{D}$. 
Here $\mathbf{f_{enc}}$ is the transformation function of the encoder in the autoencoder.
Specifically, let $\mathbf{f_{enc}}():R^d\mapsto R^p$, such that p is dimensionality of latent vector.
Also, for each $x_i$, we draw $K$ negative samples from a noise distribution $\mathcal{\hat{D}}$, denoted as $z_i \in R^d$.
Then, $z_e^i=\mathbf{f_{enc}}(z_i)$ with $z_e\in R^p$.
We inject secondary noise  $n$ to latent representation $z_e$ of each negative sample drawn from a multivariate Normal distribution, $n \sim N(0, \mathbf{I}^p)$, where ${I}^p$ is a identity covariance matrix. The noise injection is done to increase variation in negative samples, and improve model performance. 
Simplifying, we have the following.
\begin{equation}
\mathcal{L}_{est}(\hat{\theta}) = 
-\gamma  \sum_{x_i \in |\mathcal{D}|}ln(f(x_e)) 
- ln\big(1 - \frac{1}{|k|}\sum_{k}(f(z_e^i)+n)\big)
\label{eq:est2}
\end{equation}  
Minimizing this loss we can learn the parameters to capture the distribution of normal data. 
$\gamma$, a cost for mis-classifying normal data, is initially set to 1 and slowly increased to a maximum value in the third training phase. However empirically we find that results are not sensitive to $\gamma$. 

We utilize a simple two layered MLP with dropout to estimate $f(;\theta)$. It is found to work well in practice, though a more sophisticated estimator network can be designed.
The overall loss function to optimize, from Eq.~\ref{eq:loss1} and~\ref{eq:est2} is as follows.
\begin{equation}
    \mathbf{Loss} = \lambda \mathbb{1}(\mathbf{t}_r)\mathcal{L_R}   + \mathbb{1}(\mathbf{t}_e)\mathcal{L}_{est}(\hat{\theta})
    \label{eq:joint_loss}
\end{equation}
Here $\mathbb{1}(\mathbf{t})$ is an indicator variable, that can take values of 0 and 1, depending on the training phase, which will be explained in Section~\ref{sec:training}.
The training hyperparameter $\lambda$ modulates the importance of $\mathcal{L_R}$, and how it is varied is discussed in the training procedure. 
The overall architecture of our model, \textbf{C}ontrastive Learning based \textbf{H}eterogeneous \textbf{A}nomaly \textbf{D}etetctor is shown in Figure~\ref{fig:model}.
}

\subsection{Model Training}\label{sec:training}{
We make use of a three phase training procedure to learn our model.
The first phase, termed as burn-in phrase,  is training the autoencoder only, using reconstruction loss.
In this phase the estimator network is not modified. It is essential because we want to first obtain the appropriate latent representation for the data, such that normal data can be correctly reconstructed. 
Referring to Equation~\ref{eq:joint_loss}, $\mathbb{1}(\mathbf{t}_r)=1$ and $\mathbb{1}(\mathbf{t}_e)=0$.
The second phase trains both the autoencoder and the estimator jointly.
We notice that the scale of the two losses are significantly different and it is difficult to determine best weight values in an unsupervised setting.
Inspired by the training procedure in~\citeauthor{gulrajani2017improved}(\citeyear{gulrajani2017improved}), the estimator loss is included in alternate mini-batches.
That is $\mathbb{1}(\mathbf{t}_e)=1$ for alternate batches, $\mathbb{1}(\mathbf{t}_r)=1$ being constant. 
This leads to a stable joint training of the two components, avoiding requirement for a sensitive user provided scaling hyperparameter like some other models.
Referring to Eq.~\ref{eq:joint_loss}, the scaling hyperparameter $\lambda$ is set to 1 in first phase.In the second phase it is decayed using $exp(-t)$ where $t$ is the epoch number of the second phase. 
The third and final phase considers only the estimator loss, and only the estimation network is trained, keeping the encoder parameters fixed. Only $\mathbb{1}(\mathbf{t}_e)=1$ in this phase.
}

\section{Generating Negative Samples}{
Generating negative samples is key in contrastive estimation of the data distribution.
In the case of heterogeneous data, the concept of negative samples is not immediately evident as in other cases such as text or networks.
Negative samples should have adequate variation, such that they provide a contrastive background to estimate the data distribution.
They should not be entirely comprised of noise or be divergent from target distribution, but have some similarity with the normal data.
We propose a random subspace based approach applied to both the categorical and the real valued features in data. 
While unexpected combination of entities or values might be considered as negative samples, simply perturbing only the categorical features or only the continuous features might not guarantee the resulting record satisfying the aforementioned criteria. 
Further, we found simply adding noise in the latent space to latent representation of data does not provide good negative samples. Thus we adopt the algorithm which is detailed in  Algorithm~\ref{alg:neg_sam} for obtaining negative samples.

At most half of categorical features are selected, and each entity is replaced by another instance belonging to same category.
Let $a_w$ be the arity of the $w^{th}$ category. The probability of a category to be selected for perturbation is chosen following a multinomial distribution. 
The probability of $w$ being selected is the sum normalized value of $q_w$, where $q_w = (a_w/ \sum_{w`}a_{w`})^{0.75}$. 
The dampening factor $0.75$ is added following ~\cite{mikolov2013distributed}, so that not only the high arity features are perturbed.

Continuous values are assumed to be scaled with 0-1 normalization.
We add uniformly generated random noise to randomly selected  $\floor{r/2}$ features, where $r$ is the number of continuous features. 
The noise deviation parameter $\delta$ is used to shift the mean of $U(0,1)$. Intuitively using $\delta=0.5$ makes sense, and works well in our case.
The range of the noise is set up to be beyond the expected range of the features.
This per-feature perturbation forces the feature values for negative samples to spread out over and beyond the range of possible values.
}
\begin{algorithm}[tp]
\KwData{Training records $\mathcal{D}$; noise deviation $\delta$}
\KwResult{Negative samples for each training record}
\For{each training record $\mathbf{x} \in \mathcal{D}$}{
\For{i = 1:m negative samples} {
    Select $i \in [1,\floor{k/2})$ categorical features\;
    \For{each feature in $i$}{
    Replace entity with a random entity\;
    }
    Select $|j_1| = \floor{r/4}$ real features randomly\;
    \For{each feature $u \in j_1$}{
    $n \sim$ Uniform(0,1) + $\delta$, $u = u + n $
    }
    Select $|j_2| = \floor{r/4}$ real features randomly\;
    \For{each feature $v \in j_2$}{
    $n \sim$ Uniform(0,1) - $\delta$, $v = v + n $
    }
  }
}
\caption{Negative sample generation}
\label{alg:neg_sam}
\end{algorithm}
}

\section{Experiments and Results}{
Our target dataset on timber trade does not have any annotation labels regarding which records are anomalous. As such we provide a qualitative case study of the top-100 anomalies as detected by our model on the timber trade dataset. In order to showcase the efficacy of our proposed model we provide results on multiple large open source datasets with heterogeneous; further details are provided in the Appendix.
While these datasets belong to the network intrusion detection, no domain knowledge is used. 
The datasets used are:
(i) \textbf{KDDCup99}~\cite{Dua:2019}:
The KDDCup99 10 percent dataset has been a benchmark dataset for anomaly detection tasks. 
We choose the 'Normal' class label as normal data and attack classes are treated as anomalies.

(ii) \textbf{KDDCup99-N}~\cite{Dua:2019}: 
We consider the instances KDDCup99 10 percent data with attack class `Neptune' to be normal data, and data points with label 'Normal' are considered as be anomalies. 

(iii) \textbf{UNSW-NB15}~\cite{unswnb15}: 
We select data with label \textit{Normal} as normal data, and rarer attack classes \textit{Backdoor}, \textit{Analysis}, \textit{Shellcode}, \textit{Worms} as anomalies. 

(iv) \textbf{NSL-KDD}~\cite{nslkdd_dataset}: 
The combined train and test partitions of the original dataset is used, with 
`Normal' class chosen as normal data and rest as anomalies.

(v) \textbf{GureKDD}~\cite{gureKDD_dataset}: The `Normal' class label is chosen as normal data, and rest as anomalies.

\begin{table*}[ht]
    \centering
    \begin{tabular}
    {lccccccc}
    \toprule
    \textbf{Data Set} & \textbf{OCSVM}   & \textbf{FAE-r}         & \textbf{DAE} & \textbf{DCN} & \textbf{DAGMM} & \textbf{dSVDD} &\textbf{CHAD}\\ 
    \midrule
    KDDCup99    &  \makecell{\textbf{0.99}039 \\($\pm$ 0.0012)} & \makecell{0.96113 \\ ($\pm$ 0.0371)} &  \makecell{0.76297 \\($\pm$ 0.1076)} & \makecell{0.89324 \\($\pm$ 0.0447)} & \makecell{ 0.79432  \\($\pm$  0.2015)} & \makecell{  0.68164 \\ ($\pm$0.1407) } & \makecell{ 0.97332\\ ($\pm$ 0.0158)}\\
    \midrule
    KDDCup99-N  & \makecell{0.99999 \\($\pm$ 0.0001)} & \makecell{0.58391 \\ ($\pm$ 0.0002)} &  \makecell{0.99957 \\($\pm$ 0.0004)} & \makecell{0.94701 \\($\pm$ 0.0312)} & \makecell{0.99695 \\($\pm$ 0.0025)} & \makecell{ 0.97843 \\($\pm$ 0.0046)} & \makecell{ 0.99647\\ ($\pm$ 0.0035)}  \\
    \midrule
    UNSW-NB15   & \makecell{0.38445 \\($\pm$ 0.0055)} & \makecell{0.65070 \\ ($\pm$ 0.0379)} & \makecell{0.26121 \\($\pm$ 0.0498)} & \makecell{0.66786 \\($\pm$ 0.0785)} & \makecell{0.34278 \\($\pm$ 0.1980 )} & \makecell{ 0.74479 \\($\pm$ 0.1144)} & \makecell{\textbf{0.78}873 \\ ($\pm$ 0.0032)}  \\
    \midrule
    NSL-KDD     & \makecell{0.84869 \\($\pm$ 0.0059)} & \makecell{\textbf{0.88}895 \\ ($\pm$ 0.0316)} & \makecell{0.81286 \\($\pm$ 0.0329)} & \makecell{0.53152 \\($\pm$ 0.1574)} & \makecell{0.36002 \\($\pm$ 0.1440)} & \makecell{  0.60989 \\($\pm$ 0.0729)} & \makecell{ 0.75512 \\ ($\pm$ 0.0517)}  \\
    \midrule
    Gure-KDD    & \makecell{0.68652 \\($\pm$ 0.0063)} & \makecell{0.67999 \\ ($\pm$ 0.0229)} &  \makecell{0.68553 \\($\pm$ 0.0433)} & \makecell{0.37014 \\($\pm$ 0.0981)} & \makecell{0.35806 \\($\pm$ 0.0834)} & \makecell{  0.63064 \\($\pm$ 0.0719)} &\makecell{ \textbf{0.75}680 \\ ($\pm$ 0.03918)}  \\
    \midrule
    \textbf{Average Score} & 0.78201 & 0.75294 & 0.70443  & 0.68195 & 0.56960 & 0.72908 & \textbf{0.85}409\\
    \bottomrule
    \end{tabular}
    \caption{Performance comparison w.r.t to Average Precision. On average CHAD outperforms all models with an improvement of over 9.2\%  over second best model. Though CHAD is not the best in KDDCup99 and NSL-KDD datasets it does provide competitive results. For KDDCup99-N, we find there is no single definitive best performing model. Experimentally we find that unlike CHAD the other models are highly sensitive to choice of hyper-parameters (refer Fig.~\ref{fig:hyperparams}) and in absence of labelled data /domain knowledge making a good choice for hyper-parameters will be difficult. }
    \label{tab:table_results}
\end{table*}

\begin{figure*}
    \centering
    \includegraphics[width=\textwidth]{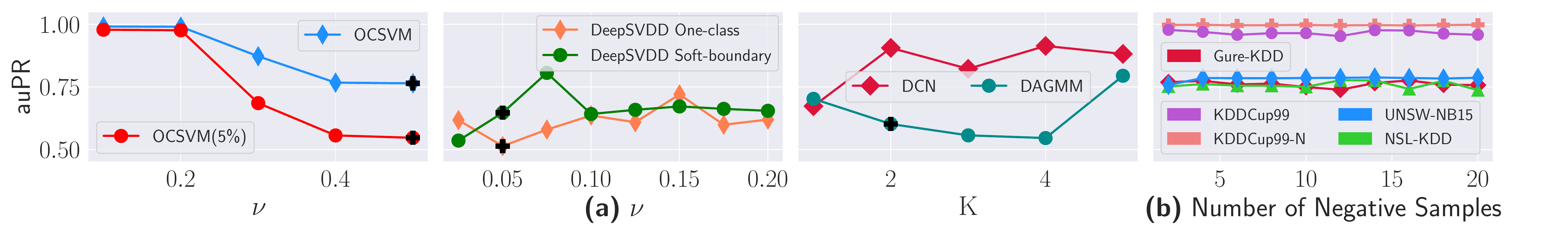}
     \caption{(a) Performance variation of competing methods to changing key hyper-parameters, for anomaly detection on KDDCup99 data. The standard or recommended hyperparameter settings are highlighted. (b) Effect of varying number of negative samples per instance on model performance for CHAD for all datasets.}
     \label{fig:hyperparams}
\end{figure*}
\subsection{Comparison with Baselines}{

To understand the effectiveness of our method, we compare the performance of our model against state-of-the-art approaches. Following previous works such as ~\citeauthor{akoglu2012fast}~(\citeyear{akoglu2012fast}) and ~\citeauthor{chen2016entity}~(\citeyear{chen2016entity}) average precision, calculated as area under Precision-Recall curve(auPR) is used as the metric. This provides a more complete picture than using F1 scores~\cite{davis2006relationship}, since the knowledge of threshold - which is the percentage of anomalies - is not assumed.
The following baselines are compared against.

\textbf{Deep SVDD}~\cite{ruff2018deep} proposed deep learning based feature extraction combined with a one class classification approach similar to OCSVM. Anomaly scores are determined as distance from the centre of the hyper-sphere, defined by points in latent space. We modify the image specific architecture to general data using 3 layered fully connected network, which is pretrained as an autoencoder. 

\textbf{DAGMM}~\cite{zong2018deep} combines a deep autoencoder with Gaussian mixture model to perform anomaly detection.
The sample likelihood as estimated by the model is used as anomaly score. The suggested 3 layered network architecture and other parameters for the KDDCup99 data set in original work is used.

\textbf{DCN}~\cite{yang2017towards} performs dimensionality reduction through a deep autoencoder and K-means clustering jointly. We use greedy layer-wise pretraining for the autoencoder to improve performance, prior to performing joint optimization. The distance of a point from its assigned cluster center is used as the anomaly score, following prior works.
We use architectures similar to DAGMM  and tune for the number of clusters and other training parameters. 

\textbf {Deep Auto Encoder}~\cite{chen2017outlier}: A deep autoencoder with dropout with greedy layer-wise pretraining is used, and reconstruction loss is used as anomaly score.
Previous works~\cite{xie2016unsupervised} point out that successive layers of dropout achieve a denoising effect. The network architecture used is similar to DCN.

\textbf {FAE-r}: We use field aware auto-encoder from our model with reconstruction loss as anomaly score. No any greedy layer-wise pretraining is used and network architecture are used is same as our complete model.

\textbf {OCSVM}~\cite{scholkopf2000support} is a classification based approach, that determines a decision boundary for normal data. We use an exponential kernel for all datasets, and find setting the hyperparameter $\nu$ to 0.1 works best.

For all models, we experiment with various choices for the respective hyper parameters and report the best score. We refer the reader to Section~\ref{subsec:hyperparams} for more details.
The experimental results are shown in Table~\ref{tab:table_results}. 
We use 10 runs, each time picking a random training and test set.
Further we use 5 sets of randomly sampled anomalies for each instance.
For our model CHAD, for each observed instance we pick 10 negative samples and optimize the model with Adam~\cite{kingma2014adam}, setting learning rate to $5 \times 10^{-4}$ and batch size 256 across all datasets. 
The number of epochs for the three training phases are set to 50,10 and 25 respectively. We use a generic 3-layered pyramid-like architecture for the autoencoder, with a dropout of 0.2. 

In judging average performance we find our model CHAD performs better than the baselines. 
OCSVM provides comparable performance in some cases.
However, it does not perform well for UNSW-NB15 dataset due to higher dimensionality caused by comparatively larger entity count.
DAE and DCN also performs well in most cases. 
DAGMM however does not perform well despite its expected good performance on general data. We observe a high variance, and it can be attributed to the fact that loss metrics augmenting the latent space does not necessarily improve model expressiveness.
FAE-r is shown to perform well in many cases, which demonstrates that our autoencoder captures the latent representation well.
However, it performs poorly on KDDCup-N, which can be attributed to over generalization issue mentioned earlier.
We find our model performs comparably or favorably in majority of cases. 
Our model does not perform the best for NSL-KDD, however in case of UNSW-NB15 and Gure-KDD it shows a strong advantage. However it is to be noted for all models like OCSVM, DCN we had selected the best hyper-parameters via a validation set, which is not easy to obtain in real-world scenarios.
In the next section we discuss about how models like OCSVM are sensitive to bad choices of hyper-parameters. 
}
\vspace{-1pt}
\subsection{Comparing the Effect of Hyperparameters}\label{subsec:hyperparams}{
Sensitive hyperparameters, which cannot be determined apriori without a validation set~\cite{oliver2018realistic}, have a significant effect on model performance.
We empirically show that competing models are highly sensitive to hyper-parameter values. The experiments are performed on the KDDCup-99 dataset where all models perform well.
Case in point, for OCSVM, although standard value of $\nu$ is 0.5 is used in available packages and in the original work, we find setting it to 0.1 obtains best results. 
Further with a smaller number of anomalies to be detected, the hyperparameter setting affects performance more drastically as shown in Figure~\ref{fig:hyperparams}(a).
In case of deepSVDD, authors recommend $\nu$ between 0.01 and 0.1, but the model performance varies significantly between those bounds, and beyond that. Moreover, while the one-class objective is shown to consistently perform better in the original work, this is not always true as shown in Figure~\ref{fig:hyperparams}(a). 
Similarly for DCN and DAGMM, the number of clusters has a significant effect on performance.
In training our model with negative samples, the count of negative samples per instance is the most significant hyper-parameter.
We train our model with varying number of negative samples per instance, and perform evaluation for all the data sets as shown shown in Figure~\ref{fig:hyperparams}(b). 
For the latter three datasets model performance is found to increase slightly with number of negative samples, though lower number of samples does not adversely affect the model performance. 
Thus our model is not sensitive to hyperparameter settings.
}

\subsection{Varying Anomaly Content}{
\begin{table}[tp]
\small
    \begin{tabular}
    {lccccc}
    \toprule
    Data Set & $2\%$ & $4\%$ & $6\%$ & $8\%$ & $10\%$ \\
    \midrule
    KDDCup99 &  0.9665  & 0.9705  & 0.9740  & 0.9739  & 0.9757\\
    KDD-N    & 0.9941 & 0.9958 & 0.9964 & 0.9967 & 0.9965\\
    NB15     & 0.7472 & 0.7630 & 0.7747 & 0.7834 & 0.7887\\
    NSL-KDD  & 0.6498 & 0.7042 & 0.7254 & 0.7401 & 0.7499\\
    Gure-KDD & 0.6905 & 0.7232 & 0.7357 & 0.7481 & 0.7560 \\
    \bottomrule
    \end{tabular}
     \caption{Average Precision score of anomaly detection task, at different percentages of anomalies in the test set.}
     \vspace{-3pt}
    \label{tab:vary_anom}
\end{table}

Given anomalies can be rare, it is important to understand how our model performs with varying percentage of anomalies. For each data set, we increase the percentages of anomalies in the combined test set from $2\%$ to $10\%$, to understand how our model performs in each scenario. 
The results are reported in Table~\ref{tab:vary_anom}. 
As expected, it is a more challenging task to detect anomalies when they are rarer; and our model performance improves slightly as the percentage of anomalies increase.
However it is also evident that CHAD is effective in cases with very low anomaly percentages.
}

\subsection{Importance of Secondary Noise}\label{sec:instance_noise}{

Adding a secondary instance noise to negative samples at the latent space of the autoencoder helps in building a better model, and our experiments provides empirical evidence that affirms this. 
This isotropic Gaussian noise helps prevent over-fitting while training the model, which  is crucial without validation sets in this unsupervised training scenario.
This is evidenced by the fact that omitting the instance noise in some training runs,  although the loss converges to a lower value the test time performance is lower. 
To better understand this, we visualize the negative samples generated for the dataset KDDCup99 with and without this noise, as shown in Figure~\ref{fig:noise_viz}.
It can be clearly seen adding noise spreads out the negative samples over data space
resulting in  a better approximation of contrastive noise distribution.
The model thus gets trained with negative samples with a significantly greater variation and is more robust. 
Table~\ref{tab:noise_table} shows the improvement in auPR provided by the addition of this secondary noise, which becomes more pronounced  when only few negative samples per instance are used.

\begin{figure}[!tp]
\centering
\includegraphics[width=0.8\columnwidth]{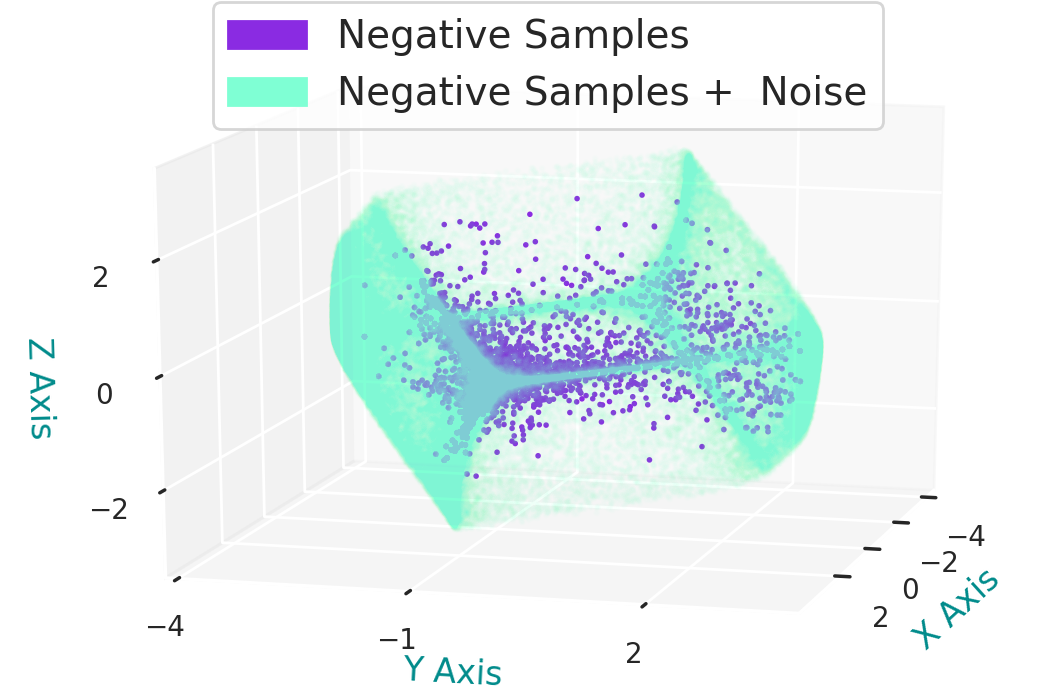}
\caption{Visualization of latent vectors of negative samples for KDDCup99 dataset, with and without injected noise.}
\label{fig:noise_viz}
\end{figure}
\begin{table}[t]
\small
    \begin{tabular}
    {p{1cm}ccccc}
    \toprule
    Secondary noise  & KDD & KDD-N & NB15 & NSL-K  &  Gure-K  \\
    \midrule
    Yes       & \textbf{0.9723} &  0.9973 & 0.7874 & \textbf{0.7682} & \textbf{0.7662} \\
    No    & 0.9325 &  0.9974 & 0.7887 & 0.6983 & 0.6943 \\
    \bottomrule
    \end{tabular}
    \caption{Average Precision score when using and not using instance noise at the autoencoder latent representation layer.}
    \label{tab:noise_table}
    \vspace{-1em}
\end{table}
}
}
\section{ Qualitative study on Timber Trade Data}{
Transaction data pertaining to international trade is extremely complex, where the challenge of detecting potentially suspicious activities is aggravated by lack of annotations.
Our target dataset of United States imports~\cite{panjiva-trade-data-2019} contains over a million rows spanning multiple years and over 70 attributes.
With the help of our collaborating domain experts, we select relevant attributes: \textit{Carrier}, \textit{ShipmentOrigin}, \textit{ShipmentDestination}, \textit{PortOfUnlading}, \textit{PortOfLading}, \textit{HSCode}, \textit{VolumeTEU}, \textit{Weight} and \textit{ValueOfGoods}.
The last three are continuous attributes, while rest are categorical. 
Harmonized System (HS) Codes describe the type of goods traded, and help incorporating domain knowledge.

We extract three datasets denoted, where training set and test set are chronologically consecutive parts of the dataset.
We filter records using HS Codes to select transactions related to timber products relevant to our use case.
Data preprocessing includes removal of very low frequency entities and records with missing values, and ensuring test set contains only previously seen entities.
Each of training sets contain around 125k records and categorical attributes have arity in order of hundreds to thousands.
We train our model CHAD on each of the three training sets, adding an extra layer to the architecture of the autoencoder while maintaining other training parameters from previous experiments.
Due to lack of labels to perform quantitative analysis, we consider the lowest scored 100 records for each test set and investigate them.
Our evaluation marks a record correct if it contains entities,values or combination with low probability; we refer to the appendix for details.
Note that we are considering only simple checks and more complex anomalies are possibly present, which can be revealed through further investigation.
For the three sets we find respectively $77\%$, $86\%$ and $91\%$ of investigated records satisfy the criteria of the preliminary tests, demonstrating our approach is effective. We discuss some examples to illustrate this. 
For the record (ID 128270910), the HS Code 440341 tells us the shipment contains timber products from hardwoods such as \textit{Meranti(Shorea spp.}. 
Neither the carrier has shipped such items previously, nor is this HS Code  seen to be from the origin (Chile).
Further such items are not seen to be transported by the carrier through the lading port or shipped to destination(PA) previously through the unlading port. 
Thus the record has multiple interesting yet unexpected patterns, and is correctly scored low by our model.
In the record (ID 130826624), we see items with HS Code 441872 originating in Slovenia, being transported through Italy which is an unobserved pattern.
Further the carrier(CGPN), is not previously observed to transport such items from lading port(Italy) or to unlading port (NY). 
Thus this record should be indeed be investigated further, and is correctly pointed out by our algorithm. 
For the record (ID 131616261) the items have a HS Code 9503, thus can contain timber products.
Given the HS Code, lading and unlading ports, the volume, weight and value have very low probability of occurring despite that none of the entities are rare.
Thus the record is aptly scored low by our model. 
Finally, let us consider the record (ID 144960415) where items with HS Code 440799, which can contain coniferous or tropical wood, are seen unexpectedly being shipped from the lading port~(Iceland). 
The lading port is not previously seen to ship from the origin (US), neither is the HS Code seen previously transported by the carrier~(EIMU) to and from the ports of lading and unlading.
Thus the record has multiple unexpected patterns, and is aptly deemed anomalous. 
These examples help demonstrate that our model is effective for the intended task.
}
\section{Conclusion and Future Work}{
Unsupervised anomaly detection in large scale real-world data is a challenging task.
The approach presented in this work tries to address some of the challenges of existing models in a systematic fashion,
and improvements are empirically demonstrated. 
We show our model CHAD, being non-parametric, can generalize to different data distributions and is not sensitive to changes in hyper-parameters.
There are however further research questions we would like to address.
This research being part of a larger project, our next steps are to explore a human-in-the-loop framework using our model.
Also currently our model does not distinguish between rare (and new / unseen instances) and unlikely combinations. 
It is important to make a difference between such instances to make the system fair and unbiased. Thus moving forward we would like to explore the angle of fairness in anomaly detection.
Further research questions pertaining to temporal aspects of trade data, model interpretability and understanding feature importance are also of consequence for our application.
Thus these questions motivate multiple continuing future research directions for us.
}
\bibliographystyle{ACM-Reference-Format}
\bibliography{sample-base}


\begin{thebibliography}{40}


\ifx \showCODEN    \undefined \def \showCODEN     #1{\unskip}     \fi
\ifx \showDOI      \undefined \def \showDOI       #1{#1}\fi
\ifx \showISBNx    \undefined \def \showISBNx     #1{\unskip}     \fi
\ifx \showISBNxiii \undefined \def \showISBNxiii  #1{\unskip}     \fi
\ifx \showISSN     \undefined \def \showISSN      #1{\unskip}     \fi
\ifx \showLCCN     \undefined \def \showLCCN      #1{\unskip}     \fi
\ifx \shownote     \undefined \def \shownote      #1{#1}          \fi
\ifx \showarticletitle \undefined \def \showarticletitle #1{#1}   \fi
\ifx \showURL      \undefined \def \showURL       {\relax}        \fi
\providecommand\bibfield[2]{#2}
\providecommand\bibinfo[2]{#2}
\providecommand\natexlab[1]{#1}
\providecommand\showeprint[2][]{arXiv:#2}

\bibitem[\protect\citeauthoryear{Abdallah, Maarof, Zainal,
  et~al\mbox{.}}{Abdallah et~al\mbox{.}}{2016}]%
        {abdallah2016fraud}
\bibfield{author}{\bibinfo{person}{Aisha Abdallah},
  \bibinfo{person}{Mohd~Aizaini Maarof}, \bibinfo{person}{Anazida Zainal},
  {et~al\mbox{.}}} \bibinfo{year}{2016}\natexlab{}.
\newblock \showarticletitle{Fraud detection system: A survey}.
\newblock \bibinfo{journal}{\emph{Journal of Network and Computer
  Applications}} (\bibinfo{year}{2016}), \bibinfo{pages}{90--113}.
\newblock


\bibitem[\protect\citeauthoryear{Akoglu, Tong, Vreeken, and Faloutsos}{Akoglu
  et~al\mbox{.}}{2012}]%
        {akoglu2012fast}
\bibfield{author}{\bibinfo{person}{Leman Akoglu}, \bibinfo{person}{Hanghang
  Tong}, \bibinfo{person}{Jilles Vreeken}, {and} \bibinfo{person}{Christos
  Faloutsos}.} \bibinfo{year}{2012}\natexlab{}.
\newblock \showarticletitle{Fast and reliable anomaly detection in categorical
  data}. In \bibinfo{booktitle}{\emph{Proceedings of the 21st ACM international
  conference on Information and knowledge management}}.
  \bibinfo{pages}{415--424}.
\newblock


\bibitem[\protect\citeauthoryear{Breunig, Kriegel, Ng, and Sander}{Breunig
  et~al\mbox{.}}{2000}]%
        {breunig2000lof}
\bibfield{author}{\bibinfo{person}{Markus~M Breunig},
  \bibinfo{person}{Hans-Peter Kriegel}, \bibinfo{person}{Raymond~T Ng}, {and}
  \bibinfo{person}{J{\"o}rg Sander}.} \bibinfo{year}{2000}\natexlab{}.
\newblock \showarticletitle{LOF: identifying density-based local outliers}. In
  \bibinfo{booktitle}{\emph{Proceedings of the 2000 ACM SIGMOD international
  conference on Management of data}}. \bibinfo{pages}{93--104}.
\newblock


\bibitem[\protect\citeauthoryear{Chalapathy and Chawla}{Chalapathy and
  Chawla}{2019}]%
        {chalapathy2019deep}
\bibfield{author}{\bibinfo{person}{Raghavendra Chalapathy} {and}
  \bibinfo{person}{Sanjay Chawla}.} \bibinfo{year}{2019}\natexlab{}.
\newblock \showarticletitle{Deep learning for anomaly detection: A survey}.
\newblock \bibinfo{journal}{\emph{arXiv preprint arXiv:1901.03407}}
  (\bibinfo{year}{2019}).
\newblock


\bibitem[\protect\citeauthoryear{Chalapathy, Menon, and Chawla}{Chalapathy
  et~al\mbox{.}}{2018}]%
        {chalapathy2018anomaly}
\bibfield{author}{\bibinfo{person}{Raghavendra Chalapathy},
  \bibinfo{person}{Aditya~Krishna Menon}, {and} \bibinfo{person}{Sanjay
  Chawla}.} \bibinfo{year}{2018}\natexlab{}.
\newblock \showarticletitle{Anomaly detection using one-class neural networks}.
\newblock \bibinfo{journal}{\emph{arXiv preprint arXiv:1802.06360}}
  (\bibinfo{year}{2018}).
\newblock


\bibitem[\protect\citeauthoryear{Chandola, Banerjee, and Kumar}{Chandola
  et~al\mbox{.}}{2009}]%
        {chandola2009anomaly}
\bibfield{author}{\bibinfo{person}{Varun Chandola}, \bibinfo{person}{Arindam
  Banerjee}, {and} \bibinfo{person}{Vipin Kumar}.}
  \bibinfo{year}{2009}\natexlab{}.
\newblock \showarticletitle{Anomaly detection: A survey}.
\newblock \bibinfo{journal}{\emph{ACM computing surveys}} \bibinfo{volume}{41},
  \bibinfo{number}{3} (\bibinfo{year}{2009}).
\newblock


\bibitem[\protect\citeauthoryear{Chen, Sathe, Aggarwal, and Turaga}{Chen
  et~al\mbox{.}}{2017}]%
        {chen2017outlier}
\bibfield{author}{\bibinfo{person}{Jinghui Chen}, \bibinfo{person}{Saket
  Sathe}, \bibinfo{person}{Charu Aggarwal}, {and} \bibinfo{person}{Deepak
  Turaga}.} \bibinfo{year}{2017}\natexlab{}.
\newblock \showarticletitle{Outlier detection with autoencoder ensembles}. In
  \bibinfo{booktitle}{\emph{SDM}}. \bibinfo{pages}{90--98}.
\newblock


\bibitem[\protect\citeauthoryear{Chen, Tang, Sun, Chen, and Zhang}{Chen
  et~al\mbox{.}}{2016}]%
        {chen2016entity}
\bibfield{author}{\bibinfo{person}{Ting Chen}, \bibinfo{person}{Lu-An Tang},
  \bibinfo{person}{Yizhou Sun}, \bibinfo{person}{Zhengzhang Chen}, {and}
  \bibinfo{person}{Kai Zhang}.} \bibinfo{year}{2016}\natexlab{}.
\newblock \showarticletitle{Entity Embedding-based Anomaly Detection for
  Heterogeneous Categorical Events}. In \bibinfo{booktitle}{\emph{IJCAI}}.
  \bibinfo{pages}{1396--1403}.
\newblock


\bibitem[\protect\citeauthoryear{Datta, Islam, Self, Meadows, Simeone,
  Outhwaite, Keong, Smith, Walker, and Ramakrishnan}{Datta
  et~al\mbox{.}}{2020}]%
        {datta2020detecting}
\bibfield{author}{\bibinfo{person}{Debanjan Datta},
  \bibinfo{person}{Mohammad~Raihanul Islam}, \bibinfo{person}{Nathan Self},
  \bibinfo{person}{Amelia Meadows}, \bibinfo{person}{John Simeone},
  \bibinfo{person}{Willow Outhwaite}, \bibinfo{person}{Chen~Hin Keong},
  \bibinfo{person}{Amy Smith}, \bibinfo{person}{Linda Walker}, {and}
  \bibinfo{person}{Naren Ramakrishnan}.} \bibinfo{year}{2020}\natexlab{}.
\newblock \showarticletitle{Detecting Suspicious Timber Trades.}. In
  \bibinfo{booktitle}{\emph{Thirty Second IAAI Conference}}.
\newblock


\bibitem[\protect\citeauthoryear{Davis and Goadrich}{Davis and
  Goadrich}{2006}]%
        {davis2006relationship}
\bibfield{author}{\bibinfo{person}{Jesse Davis} {and} \bibinfo{person}{Mark
  Goadrich}.} \bibinfo{year}{2006}\natexlab{}.
\newblock \showarticletitle{The relationship between Precision-Recall and ROC
  curves}. In \bibinfo{booktitle}{\emph{Proceedings of the 23rd international
  conference on Machine learning}}. \bibinfo{pages}{233--240}.
\newblock


\bibitem[\protect\citeauthoryear{Dua and Graff}{Dua and Graff}{2017}]%
        {Dua:2019}
\bibfield{author}{\bibinfo{person}{Dheeru Dua} {and} \bibinfo{person}{Casey
  Graff}.} \bibinfo{year}{2017}\natexlab{}.
\newblock \bibinfo{title}{{UCI} Machine Learning Repository}.
\newblock
\newblock
\urldef\tempurl%
\url{http://archive.ics.uci.edu/ml}
\showURL{%
\tempurl}


\bibitem[\protect\citeauthoryear{Goldstein and Uchida}{Goldstein and
  Uchida}{2016}]%
        {goldstein2016comparative}
\bibfield{author}{\bibinfo{person}{M. Goldstein} {and} \bibinfo{person}{S.
  Uchida}.} \bibinfo{year}{2016}\natexlab{}.
\newblock \showarticletitle{A comparative evaluation of unsupervised anomaly
  detection algorithms for multivariate data}.
\newblock \bibinfo{journal}{\emph{PloS one}} \bibinfo{volume}{11},
  \bibinfo{number}{4} (\bibinfo{year}{2016}).
\newblock


\bibitem[\protect\citeauthoryear{Gulrajani, Ahmed, Arjovsky, Dumoulin, and
  Courville}{Gulrajani et~al\mbox{.}}{2017}]%
        {gulrajani2017improved}
\bibfield{author}{\bibinfo{person}{Ishaan Gulrajani}, \bibinfo{person}{Faruk
  Ahmed}, \bibinfo{person}{Martin Arjovsky}, \bibinfo{person}{Vincent
  Dumoulin}, {and} \bibinfo{person}{Aaron~C Courville}.}
  \bibinfo{year}{2017}\natexlab{}.
\newblock \showarticletitle{Improved training of wasserstein gans}. In
  \bibinfo{booktitle}{\emph{Advances in neural information processing
  systems}}. \bibinfo{pages}{5767--5777}.
\newblock


\bibitem[\protect\citeauthoryear{Gutmann and Hyv{\"a}rinen}{Gutmann and
  Hyv{\"a}rinen}{2010}]%
        {gutmann2010noise}
\bibfield{author}{\bibinfo{person}{Michael Gutmann} {and} \bibinfo{person}{Aapo
  Hyv{\"a}rinen}.} \bibinfo{year}{2010}\natexlab{}.
\newblock \showarticletitle{Noise-contrastive estimation: A new estimation
  principle for unnormalized statistical models}. In
  \bibinfo{booktitle}{\emph{Proceedings of the Thirteenth International
  Conference on Artificial Intelligence and Statistics}}.
  \bibinfo{pages}{297--304}.
\newblock


\bibitem[\protect\citeauthoryear{Hinton and Salakhutdinov}{Hinton and
  Salakhutdinov}{2006}]%
        {hinton2006reducing}
\bibfield{author}{\bibinfo{person}{Geoffrey~E Hinton} {and}
  \bibinfo{person}{Ruslan~R Salakhutdinov}.} \bibinfo{year}{2006}\natexlab{}.
\newblock \showarticletitle{Reducing the dimensionality of data with neural
  networks}.
\newblock \bibinfo{journal}{\emph{science}} \bibinfo{volume}{313},
  \bibinfo{number}{5786} (\bibinfo{year}{2006}), \bibinfo{pages}{504--507}.
\newblock


\bibitem[\protect\citeauthoryear{Khraisat, Gondal, Vamplew, and
  Kamruzzaman}{Khraisat et~al\mbox{.}}{2019}]%
        {khraisat2019survey}
\bibfield{author}{\bibinfo{person}{Ansam Khraisat}, \bibinfo{person}{Iqbal
  Gondal}, \bibinfo{person}{Peter Vamplew}, {and} \bibinfo{person}{Joarder
  Kamruzzaman}.} \bibinfo{year}{2019}\natexlab{}.
\newblock \showarticletitle{Survey of intrusion detection systems: techniques,
  datasets and challenges}.
\newblock \bibinfo{journal}{\emph{Cybersecurity}} \bibinfo{volume}{2},
  \bibinfo{number}{1} (\bibinfo{year}{2019}), \bibinfo{pages}{20}.
\newblock


\bibitem[\protect\citeauthoryear{Kingma and Ba}{Kingma and Ba}{2015}]%
        {kingma2014adam}
\bibfield{author}{\bibinfo{person}{Diederik~P Kingma} {and}
  \bibinfo{person}{Jimmy Ba}.} \bibinfo{year}{2015}\natexlab{}.
\newblock \showarticletitle{Adam: A method for stochastic optimization}.
\newblock  (\bibinfo{year}{2015}).
\newblock


\bibitem[\protect\citeauthoryear{Latecki, Lazarevic, and Pokrajac}{Latecki
  et~al\mbox{.}}{2007}]%
        {latecki2007outlier}
\bibfield{author}{\bibinfo{person}{Longin~Jan Latecki},
  \bibinfo{person}{Aleksandar Lazarevic}, {and} \bibinfo{person}{Dragoljub
  Pokrajac}.} \bibinfo{year}{2007}\natexlab{}.
\newblock \showarticletitle{Outlier detection with kernel density functions}.
  In \bibinfo{booktitle}{\emph{International Workshop on Machine Learning and
  Data Mining in Pattern Recognition}}. Springer, \bibinfo{pages}{61--75}.
\newblock


\bibitem[\protect\citeauthoryear{Lawson and MacFaul}{Lawson and
  MacFaul}{2010}]%
        {lawson2010illegal}
\bibfield{author}{\bibinfo{person}{Sam Lawson} {and} \bibinfo{person}{Larry
  MacFaul}.} \bibinfo{year}{2010}\natexlab{}.
\newblock \showarticletitle{Illegal logging and related trade}.
\newblock  (\bibinfo{year}{2010}).
\newblock


\bibitem[\protect\citeauthoryear{Liu, Ting, and Zhou}{Liu
  et~al\mbox{.}}{2008}]%
        {liu2008isolation}
\bibfield{author}{\bibinfo{person}{Fei~Tony Liu}, \bibinfo{person}{Kai~Ming
  Ting}, {and} \bibinfo{person}{Zhi-Hua Zhou}.}
  \bibinfo{year}{2008}\natexlab{}.
\newblock \showarticletitle{Isolation forest}. In
  \bibinfo{booktitle}{\emph{2008 Eighth IEEE International Conference on Data
  Mining}}. IEEE, \bibinfo{pages}{413--422}.
\newblock


\bibitem[\protect\citeauthoryear{Mikolov, Sutskever, Chen, Corrado, and
  Dean}{Mikolov et~al\mbox{.}}{2013}]%
        {mikolov2013distributed}
\bibfield{author}{\bibinfo{person}{Tomas Mikolov}, \bibinfo{person}{Ilya
  Sutskever}, \bibinfo{person}{Kai Chen}, \bibinfo{person}{Greg~S Corrado},
  {and} \bibinfo{person}{Jeff Dean}.} \bibinfo{year}{2013}\natexlab{}.
\newblock \showarticletitle{Distributed representations of words and phrases
  and their compositionality}. In \bibinfo{booktitle}{\emph{Advances in neural
  information processing systems}}. \bibinfo{pages}{3111--3119}.
\newblock


\bibitem[\protect\citeauthoryear{{Moustafa} and {Slay}}{{Moustafa} and
  {Slay}}{2015}]%
        {unswnb15}
\bibfield{author}{\bibinfo{person}{N. {Moustafa}} {and} \bibinfo{person}{J.
  {Slay}}.} \bibinfo{year}{2015}\natexlab{}.
\newblock \showarticletitle{UNSW-NB15: a comprehensive data set for network
  intrusion detection systems (UNSW-NB15 network data set)}. In
  \bibinfo{booktitle}{\emph{2015 Military Communications and Information
  Systems Conference (MilCIS)}}. \bibinfo{pages}{1--6}.
\newblock


\bibitem[\protect\citeauthoryear{Oliver, Odena, Raffel, Cubuk, and
  Goodfellow}{Oliver et~al\mbox{.}}{2018}]%
        {oliver2018realistic}
\bibfield{author}{\bibinfo{person}{Avital Oliver}, \bibinfo{person}{Augustus
  Odena}, \bibinfo{person}{Colin~A Raffel}, \bibinfo{person}{Ekin~Dogus Cubuk},
  {and} \bibinfo{person}{Ian Goodfellow}.} \bibinfo{year}{2018}\natexlab{}.
\newblock \showarticletitle{Realistic evaluation of deep semi-supervised
  learning algorithms}. In \bibinfo{booktitle}{\emph{Advances in neural
  information processing systems}}. \bibinfo{pages}{3235--3246}.
\newblock


\bibitem[\protect\citeauthoryear{Panjiva}{Panjiva}{2019}]%
        {panjiva-trade-data-2019}
\bibfield{author}{\bibinfo{person}{Panjiva}.} \bibinfo{year}{2019}\natexlab{}.
\newblock \bibinfo{title}{Panjiva Trade Data}.
\newblock \bibinfo{howpublished}{\url{https://panjiva.com}}.
\newblock


\bibitem[\protect\citeauthoryear{Pelleg, Moore, et~al\mbox{.}}{Pelleg
  et~al\mbox{.}}{2000}]%
        {pelleg2000x}
\bibfield{author}{\bibinfo{person}{Dan Pelleg}, \bibinfo{person}{Andrew~W
  Moore}, {et~al\mbox{.}}} \bibinfo{year}{2000}\natexlab{}.
\newblock \showarticletitle{X-means: Extending k-means with efficient
  estimation of the number of clusters.}. In \bibinfo{booktitle}{\emph{Icml}},
  Vol.~\bibinfo{volume}{1}. \bibinfo{pages}{727--734}.
\newblock


\bibitem[\protect\citeauthoryear{Perona, Gurrutxaga, Arbelaitz, Mart\'{\i}n,
  Muguerza, and P\'{e}rez}{Perona et~al\mbox{.}}{2008}]%
        {gureKDD_dataset}
\bibfield{author}{\bibinfo{person}{I\~{n}igo Perona}, \bibinfo{person}{Ibai
  Gurrutxaga}, \bibinfo{person}{Olatz Arbelaitz}, \bibinfo{person}{Jos\'{e}~I.
  Mart\'{\i}n}, \bibinfo{person}{Javier Muguerza}, {and}
  \bibinfo{person}{Jes\'{u}s~Ma P\'{e}rez}.} \bibinfo{year}{2008}\natexlab{}.
\newblock \showarticletitle{Service-Independent Payload Analysis to Improve
  Intrusion Detection in Network Traffic}. In
  \bibinfo{booktitle}{\emph{Proceedings of the 7th Australasian Data Mining
  Conference}}. \bibinfo{publisher}{Australian Computer Society, Inc.},
  \bibinfo{address}{AUS}.
\newblock


\bibitem[\protect\citeauthoryear{Rasmussen}{Rasmussen}{2000}]%
        {rasmussen2000infinite}
\bibfield{author}{\bibinfo{person}{Carl~Edward Rasmussen}.}
  \bibinfo{year}{2000}\natexlab{}.
\newblock \showarticletitle{The infinite Gaussian mixture model}. In
  \bibinfo{booktitle}{\emph{Advances in neural information processing
  systems}}. \bibinfo{pages}{554--560}.
\newblock


\bibitem[\protect\citeauthoryear{Ruff, Vandermeulen, Goernitz, Deecke,
  Siddiqui, Binder, M{\"u}ller, and Kloft}{Ruff et~al\mbox{.}}{2018}]%
        {ruff2018deep}
\bibfield{author}{\bibinfo{person}{Lukas Ruff}, \bibinfo{person}{Robert
  Vandermeulen}, \bibinfo{person}{Nico Goernitz}, \bibinfo{person}{Lucas
  Deecke}, \bibinfo{person}{Shoaib~Ahmed Siddiqui}, \bibinfo{person}{Alexander
  Binder}, \bibinfo{person}{Emmanuel M{\"u}ller}, {and} \bibinfo{person}{Marius
  Kloft}.} \bibinfo{year}{2018}\natexlab{}.
\newblock \showarticletitle{Deep one-class classification}. In
  \bibinfo{booktitle}{\emph{International conference on machine learning}}.
  \bibinfo{pages}{4393--4402}.
\newblock


\bibitem[\protect\citeauthoryear{Sakurada and Yairi}{Sakurada and
  Yairi}{2014}]%
        {sakurada2014ae}
\bibfield{author}{\bibinfo{person}{Mayu Sakurada} {and}
  \bibinfo{person}{Takehisa Yairi}.} \bibinfo{year}{2014}\natexlab{}.
\newblock \showarticletitle{Anomaly Detection Using Autoencoders with Nonlinear
  Dimensionality Reduction} \emph{(\bibinfo{series}{MLSDA'14})}.
  \bibinfo{publisher}{Association for Computing Machinery},
  \bibinfo{address}{New York, NY, USA}, \bibinfo{pages}{4–11}.
\newblock
\showISBNx{9781450331593}
\urldef\tempurl%
\url{https://doi.org/10.1145/2689746.2689747}
\showDOI{\tempurl}


\bibitem[\protect\citeauthoryear{Sch{\"o}lkopf, Williamson, Smola,
  Shawe-Taylor, and Platt}{Sch{\"o}lkopf et~al\mbox{.}}{2000}]%
        {scholkopf2000support}
\bibfield{author}{\bibinfo{person}{Bernhard Sch{\"o}lkopf},
  \bibinfo{person}{Robert~C Williamson}, \bibinfo{person}{Alex~J Smola},
  \bibinfo{person}{John Shawe-Taylor}, {and} \bibinfo{person}{John~C Platt}.}
  \bibinfo{year}{2000}\natexlab{}.
\newblock \showarticletitle{Support vector method for novelty detection}. In
  \bibinfo{booktitle}{\emph{Advances in neural information processing
  systems}}. \bibinfo{pages}{582--588}.
\newblock


\bibitem[\protect\citeauthoryear{Spigler}{Spigler}{2019}]%
        {spigler2019denoising}
\bibfield{author}{\bibinfo{person}{Giacomo Spigler}.}
  \bibinfo{year}{2019}\natexlab{}.
\newblock \showarticletitle{Denoising autoencoders for overgeneralization in
  neural networks}.
\newblock \bibinfo{journal}{\emph{IEEE Transactions on Pattern Analysis and
  Machine Intelligence}} \bibinfo{volume}{42}, \bibinfo{number}{4}
  (\bibinfo{year}{2019}), \bibinfo{pages}{998--1004}.
\newblock


\bibitem[\protect\citeauthoryear{Tagawa, Tadokoro, Yairi, et~al\mbox{.}}{Tagawa
  et~al\mbox{.}}{2015}]%
        {tagawa2015structured}
\bibfield{author}{\bibinfo{person}{Takaaki Tagawa}, \bibinfo{person}{Yukihiro
  Tadokoro}, \bibinfo{person}{Takehisa Yairi}, {et~al\mbox{.}}}
  \bibinfo{year}{2015}\natexlab{}.
\newblock \showarticletitle{Structured denoising autoencoder for fault
  detection and analysis}. In \bibinfo{booktitle}{\emph{Asian Conference on
  Machine Learning}}. \bibinfo{pages}{96--111}.
\newblock


\bibitem[\protect\citeauthoryear{Tavallaee, Bagheri, Lu, and
  Ghorbani}{Tavallaee et~al\mbox{.}}{2009}]%
        {nslkdd_dataset}
\bibfield{author}{\bibinfo{person}{Mahbod Tavallaee}, \bibinfo{person}{Ebrahim
  Bagheri}, \bibinfo{person}{Wei Lu}, {and} \bibinfo{person}{Ali~A Ghorbani}.}
  \bibinfo{year}{2009}\natexlab{}.
\newblock \showarticletitle{A detailed analysis of the KDD CUP 99 data set}. In
  \bibinfo{booktitle}{\emph{2009 IEEE symposium on computational intelligence
  for security and defense applications}}. IEEE, \bibinfo{pages}{1--6}.
\newblock


\bibitem[\protect\citeauthoryear{Tax and Duin}{Tax and Duin}{2004}]%
        {tax2004support}
\bibfield{author}{\bibinfo{person}{David~MJ Tax} {and}
  \bibinfo{person}{Robert~PW Duin}.} \bibinfo{year}{2004}\natexlab{}.
\newblock \showarticletitle{Support vector data description}.
\newblock \bibinfo{journal}{\emph{Machine learning}} \bibinfo{volume}{54},
  \bibinfo{number}{1} (\bibinfo{year}{2004}), \bibinfo{pages}{45--66}.
\newblock


\bibitem[\protect\citeauthoryear{Wiedenhoeft, Simeone, Smith,
  et~al\mbox{.}}{Wiedenhoeft et~al\mbox{.}}{2019}]%
        {wiedenhoeft2019fraud}
\bibfield{author}{\bibinfo{person}{A.~C. Wiedenhoeft}, \bibinfo{person}{J.
  Simeone}, \bibinfo{person}{A. Smith}, {et~al\mbox{.}}}
  \bibinfo{year}{2019}\natexlab{}.
\newblock \showarticletitle{{Fraud and misrepresentation in retail forest
  products exceeds US forensic wood science capacity}}.
\newblock \bibinfo{journal}{\emph{PloS one}} \bibinfo{volume}{14},
  \bibinfo{number}{7} (\bibinfo{year}{2019}).
\newblock


\bibitem[\protect\citeauthoryear{Xie, Girshick, Farhadi, et~al\mbox{.}}{Xie
  et~al\mbox{.}}{2016}]%
        {xie2016unsupervised}
\bibfield{author}{\bibinfo{person}{Junyuan Xie}, \bibinfo{person}{Ross
  Girshick}, \bibinfo{person}{Ali Farhadi}, {et~al\mbox{.}}}
  \bibinfo{year}{2016}\natexlab{}.
\newblock \showarticletitle{Unsupervised deep embedding for clustering
  analysis}. In \bibinfo{booktitle}{\emph{International conference on machine
  learning}}. \bibinfo{pages}{478--487}.
\newblock


\bibitem[\protect\citeauthoryear{Xu, Caramanis, and Sanghavi}{Xu
  et~al\mbox{.}}{2010}]%
        {xu2010robust}
\bibfield{author}{\bibinfo{person}{Huan Xu}, \bibinfo{person}{Constantine
  Caramanis}, {and} \bibinfo{person}{Sujay Sanghavi}.}
  \bibinfo{year}{2010}\natexlab{}.
\newblock \showarticletitle{Robust PCA via outlier pursuit}. In
  \bibinfo{booktitle}{\emph{Advances in neural information processing
  systems}}. \bibinfo{pages}{2496--2504}.
\newblock


\bibitem[\protect\citeauthoryear{Yang, Fu, Sidiropoulos, and Hong}{Yang
  et~al\mbox{.}}{2017}]%
        {yang2017towards}
\bibfield{author}{\bibinfo{person}{Bo Yang}, \bibinfo{person}{Xiao Fu},
  \bibinfo{person}{Nicholas~D Sidiropoulos}, {and} \bibinfo{person}{Mingyi
  Hong}.} \bibinfo{year}{2017}\natexlab{}.
\newblock \showarticletitle{Towards k-means-friendly spaces: Simultaneous deep
  learning and clustering}. In \bibinfo{booktitle}{\emph{International
  Conference on Machine Learning}}. \bibinfo{pages}{3861--3870}.
\newblock


\bibitem[\protect\citeauthoryear{Zhou and Paffenroth}{Zhou and
  Paffenroth}{2017}]%
        {zhou2017anomaly}
\bibfield{author}{\bibinfo{person}{Chong Zhou} {and} \bibinfo{person}{Randy~C
  Paffenroth}.} \bibinfo{year}{2017}\natexlab{}.
\newblock \showarticletitle{Anomaly detection with robust deep autoencoders}.
  In \bibinfo{booktitle}{\emph{Proceedings of the 23rd ACM SIGKDD International
  Conference on Knowledge Discovery and Data Mining}}.
  \bibinfo{pages}{665--674}.
\newblock


\bibitem[\protect\citeauthoryear{Zong, Song, Min, Cheng, Lumezanu, Cho, and
  Chen}{Zong et~al\mbox{.}}{2018}]%
        {zong2018deep}
\bibfield{author}{\bibinfo{person}{Bo Zong}, \bibinfo{person}{Qi Song},
  \bibinfo{person}{Martin~Renqiang Min}, \bibinfo{person}{Wei Cheng},
  \bibinfo{person}{Cristian Lumezanu}, \bibinfo{person}{Daeki Cho}, {and}
  \bibinfo{person}{Haifeng Chen}.} \bibinfo{year}{2018}\natexlab{}.
\newblock \showarticletitle{Deep autoencoding gaussian mixture model for
  unsupervised anomaly detection}. In \bibinfo{booktitle}{\emph{International
  Conference on Learning Representations}}.
\newblock


\end{thebibliography}

\newpage
\appendix
\section{Appendix}{
\subsection{Data Description}{
The details of the various datasets used in our experiments are provided in Table.~\ref{tab:table_datasets}. All these data sets are open source belonging to network intrusion domain, however no domain knowledge is utilized.
\begin{table}[hp]
    \centering
    \small
    \begin{tabular}
    {lccccc}
    \toprule
    \textbf{Data Set} & \makecell{\textbf{\#Cat.}}  &   \makecell{\textbf{\#Real}}   & \textbf{Train} & \textbf{Test} & \makecell{\textbf{Anom.}\\\textbf{Ratio}}\\ 
    \midrule
    \makecell{KDDCup99\\}   &  6(85) &  35 &  68094  & 29183 & 1:5 \\
    \makecell{KDDCup99N\\}  &  6(81) &  34 &  75040  & 32160 & 1:5 \\
    \makecell{UNSW-NB15\\}   &  5(158) &  37 &  39200  & 16800 & 1:10 \\
    \makecell{NSL-KDD\\}     &  7(92)  & 36 &  48646  & 20848 & 1:10 \\
    \makecell{ Gure-KDD\\}    &  8(48)  & 33 &  70000  & 30000 & 1:10 \\
    \bottomrule
    \end{tabular}
    \caption{Description of datasets used for experiments. For categorical attributes total entities are placed in parenthesis.}
    \label{tab:table_datasets}
    
\end{table}
}
\vspace{-3.5em}
\subsection{DAGMM Results}{
One of the baselines used is Deep Autoencoding Gaussian Mixture Model For Unsupervised Anomaly Detection(DAGMM) (Zong. et al.(2018)).
We observe a significant amount of variation in results across different runs, and is also found to be sensitive to different settings of hyperparameter as  previously discussed.
Though Zong et al. (2018) report state-of-art results in terms for F-1 score for KDDCup-99, the same is not seen when using Average precision as the metric. Thus in order to show that we performed faithful comparison, we show the precision, recall and F1 results for KDDCup99 dataset for DAGMM as obtained in our runs.
We find that while some of the runs achieve the reported results, as shown in Table~\ref{tab:dagmm2}, the performance is not consistent. 
\begin{table}[!hp]
    \centering
    \small
    \begin{tabular}
    {lcc}
    \toprule
      & F1-score(K=2) & F1-score(K=5)\\
    \midrule
   
    Run 1  & 0.7548          &\textbf{0.8874} \\
    Run 2  & \textbf{0.8764} &         0.3671 \\
    Run 3  & 0.6333          &\textbf{0.9659} \\
    Run 4  & 0.7136          &         0.7213 \\
    Run 5  & 0.6655          &\textbf{0.8992} \\
    Run 6  & 0.7985          &         0.6884 \\
    Run 7  & 0.6665          &\textbf{0.9353} \\
    Run 8  & 0.7233          &         0.6732 \\
    Run 9  & 0.6996          &         0.8956 \\ 
    Run 10 & \textbf{0.9350} &         0.7235 \\
    \midrule
    Mean   & 0.7467          &         0.7757 \\
    
    \bottomrule
    \end{tabular}
\caption{F1-scores for a set of 10 runs. It is observed the F1-scores reported by the original work are seen on some runs, it is not observed consistently and there is significant variance. The F1-scores that are approximately close to the reported values have been highlighted.
The mean values for precision and recall are  0.7074, 0.7996 for K=2, and 0.7585, 0.8003 for K=5 respectively. K is the number of GMM components.
}
\label{tab:dagmm2}
\vspace{-2em}
\end{table}
}

\subsection{Further Details on the Qualitative Study}{
The trade data set of interest to us is from Panjiva Inc.
The data for shipments pertaining to US import is of the period 2015 to 2017, with individual records dated by date of arrival at unlading port. 
Such data is obtained from the Bill of Lading (shipping or cargo manifests) which accompanies each import. 
A crucial component in this data is Harmonized Tariff Schedule Codes or HS Codes, which are utilized by law enforcement and customs for monitoring purposes. 
These are 10 digit codes with a staggered granularity. 
The first six digits are standard across the world, while the last four are country specific. We use the first two and four digits, to filter trade records by HS Codes.
This is a high arity attribute, for instance in first dataset the arity is 1545.

The qualitative study we perform performs a conceptually simple evaluation as a first step.
This evaluation is comprised of two steps.
First we check all entity, pair of entities or a triplet of entities that are present in test records which are highlighted by the algorithm.
If an entity is in the lowest quartile in terms of frequency, among other entities in the domain - that record is marked as a positive, since frequency distribution is highly non-uniform (skewed).
Also combinations of entities that are in the lowest quartile in terms of frequency, which includes an unforeseen co-occurrence with respect to the training data - are also marked as positive outcome.
The second part looks at the combination of numerical and categorical columns.
Specifically, given a HS Code, and an entity or a pair of entities, for each numerical column the mean and standard deviation are calculated, provided significant number of such records (over 50) exist. For test records where the values of the numerical columns do not fall within a two standard deviation of of the mean, have values which are surprising or unexpected - and thus those records are marked positive as well. These cumulative sum is tallied against total number of records considered to calculate the "hit rate". This is not a complete evaluation and more details can be uncovered through further inputs of domain experts, which is part of our ongoing work. But this study demonstrates the efficacy of our approach.
}

\subsection{Programming environment}{
For all our experiments we use Python 3.7, running on Ubuntu 18.04. The deep learning library used is PyTorch 1.4, along with  Scikit-Learn, NumPy and other standard libraries. 
The hardware used is multi-core CPU and GPU (Nvidia Tesla P100 16GB) for accelerated training. 
}
\subsection{Hyperparameter settings}{
For all data sets, our model CHAD uses the following network architectures and hyperparameters. 
\begin{itemize}
    \item Network architecture: For KDDCup99 and KDDCup99-N the network layer sizes are (64,32,16). For UNSW-NB15, NSL-KDD and Gure-KDD the layer sizes are (48,24,12), (32,16,12), (48,24,12) respectively. 
    \item Autoencoder dropout: 0.2; Scoring MLP dropout: 0.1
    \item Scoring MLP network architecture used is ($l$,$\floor{l/2}$,1), with activation function tanh in the intermediate layer and sigmoid for final output. Here $l$ is the dimensionality of the latent vector obtained from the autoencoder.
    \item  Batch size: 256; Learning rate: $5\times 10^{-4}$
    
\end{itemize}
}

}
\end{document}